\documentclass[letterpaper, 10 pt, conference]{ieeeconf}

\IEEEoverridecommandlockouts
\usepackage{amsmath}
\usepackage{amssymb}
\usepackage{graphicx}
\usepackage{booktabs}
\usepackage{subcaption}
\usepackage{caption}  
\usepackage{algorithm}
\usepackage{algorithmic}
\usepackage{multirow}
\usepackage{wrapfig}
\usepackage{comment}
\usepackage{stfloats}  
\usepackage{hyperref}

\title{\LARGE \bf
Invariance Co-training for Robot Visual Generalization
}

\author{Jonathan Yang, Chelsea Finn, Dorsa Sadigh
\thanks{}
\thanks{}%
}

\begin{document}

\maketitle

\begin{figure*}
\includegraphics[width=\textwidth]{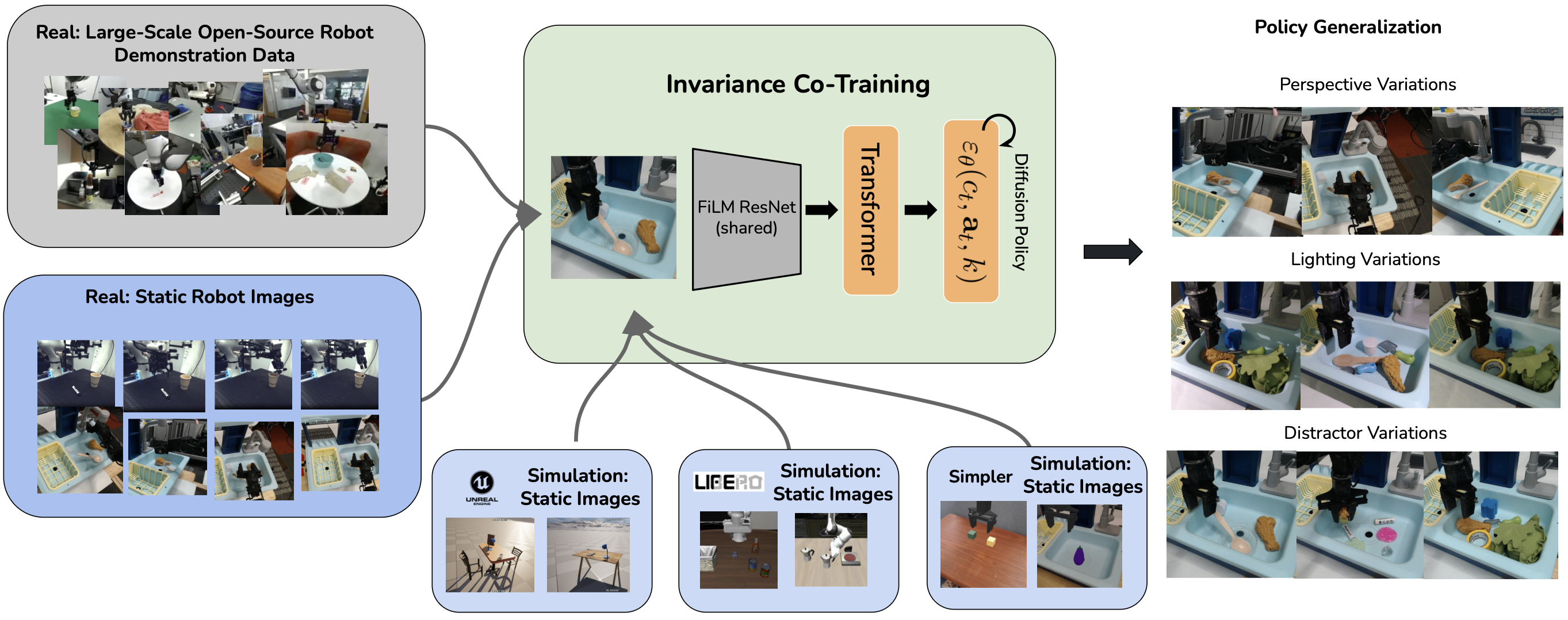}

\captionof{figure}{\textbf{Invariance Co-training.} Our method leverages diverse synthetic images, large-scale open-source datasets, and videos of static scenes to train a 2D vision encoder that generalizes to new camera viewpoints, lighting conditions, and background clutter. This encoder is co-trained with robot demonstration data to enable the policy to generalize effectively to novel visual conditions.}
\label{fig:teaser}
\end{figure*}


\begin{abstract}
Reasoning from diverse observations is a fundamental capability for generalist robot policies to operate in a wide range of environments. Despite recent advancements, many large-scale robotic policies still remain sensitive to key sources of observational variation—such as changes in camera perspective, lighting, and the presence of distractor objects. We posit that the limited generalizability of these models arises from the substantial diversity required to robustly cover these quasistatic axes, coupled with the current scarcity of large-scale robotic datasets that exhibit rich variation across them. In this work, we propose to systematically examine what robots need to generalize across these challenging axes by introducing two key auxiliary tasks—state similarity and invariance to observational perturbations—applied to both demonstration data and static visual data. We then show that via these auxiliary tasks, leveraging both more-expensive robotic demonstration data and less-expensive, visually rich synthetic images generated from non-physics-based simulation (e.g., Unreal Engine) can lead to substantial increases in generalization to unseen camera viewpoints, lighting configurations, and distractor conditions. Our results demonstrate that co-training on this diverse data improves performance by $18\%$ over existing generative augmentation methods. For more information and videos, please visit \href{https://invariance-cotraining.github.io/}{https://invariance-cotraining.github.io/}.
\end{abstract}

\section{Introduction}

Robotic foundation models have shown impressive progress in generalizing to everyday scenarios by leveraging large-scale datasets spanning multiple embodiments, environments, and tasks \cite{black2024pi0visionlanguageactionflowmodel, geminiroboticsteam2025geminiroboticsbringingai}. However, despite their breadth, the resulting models often remain brittle in real-world settings—failing to handle unseen spatial configurations of objects or adapt to drastic visual changes such as lighting and viewpoint shifts. We hypothesize that the brittleness of current robotic policies stems from insufficient coverage of key observational factors during training. For example, many large-scale datasets provide only one or two third-person perspectives per scene, limiting robustness to viewpoint shifts. Similarly, combinatorially covering all possible spatial configurations of objects is difficult in purely real-world datasets. 

Prior work has attempted to solve this coverage issue by leveraging simulation as a low-cost alternative to expensive real-world data collection. Such efforts include automating the generation of assets, scenes, and tasks within simulation \cite{nasiriany2024robocasalargescalesimulationeveryday} and developing methods for bringing real-world data into simulation \cite{torne2024reconcilingrealitysimulationrealtosimtoreal}. Despite these advances, creating diverse, high-fidelity simulated environments is expensive, particularly when each environment must have both accurate physics and visuals. Our key insight is that many difficult axes of real-world generalization in robotics can be effectively addressed using only static image data, by learning representations invariant to perturbations in robotic observation given a (state, goal) pair. Robots can learn the desired state-goal representations efficiently using diverse observations generated from quasistatic scenes, rather than physical simulations. By focusing on quasi-static scenes, simulation data can be scaled more easily to cover critical visual variations without the complexity and computational burden of modeling realistic dynamics. In this work, we investigate how co-training robot policies with large-scale, visually diverse synthetic data can improve generalization along three important axes: camera perspective, distractor clutter, and lighting variation.

How can we learn the desired state-goal invariances from static or quasi-static data? We identify two crucial capabilities for the learned representation: state similarity recognition, which involves mapping varied observation configurations (differing in viewpoint, lighting, etc.) to the same underlying state, and relative observation disentanglement, which involves understanding whether a change in observation was induced by a change in state or a (state, goal)-invariant observation perturbation. To learn representations exhibiting these properties, we employ a contrastive learning framework that encourages invariant embeddings, and utilizes our diverse quasi-static dataset to provide the necessary examples of observational variations. To further ground the learned representation and ensure it focuses on relevant physical entities, we supplement this framework with auxiliary supervised learning losses, such as predicting bounding boxes for key objects that define the task state.

Our main contribution is a scalable method for aligning robot policy encoder representations to improve generalization across multiple observational axes—specifically camera viewpoint, lighting variation, and distractor clutter—without relying on accurate physical simulation, which we coin as \emph{invariance co-training}. Unlike prior work that focuses on precise dynamics modeling or task-specific data augmentation, we demonstrate that learning structured invariances from synthetic, visually diverse images alone is sufficient to significantly improve generalization. Our method trains vision encoders on diverse synthetic data, combined with real videos of static scenes and real robot demonstration data, to enable real robot policies to generalize more broadly. Experimentally, our invariance co-training approach significantly outperforms baseline Behavioral Cloning (\textit{BC}), improving average success rates by approximately $40\%$ across key variations. Furthermore, it yields $18\%$ higher success rates compared to variants relying only on simulation or generative models.

\section{Related Work}
\label{sec:citations}

\noindent \textbf{Sim-to-real Transfer.} 
Sim-to-real transfer harnesses simulation as a cost-effective means to supply training data for robotic policies. However, mismatches in visual appearance and physics accuracy can create a sim-to-real gap that can lead to difficulty in utilizing this data. To solve this issue, prior works have either utilized domain randomization \cite{tobin2017domain, tremblay2018trainingdeepnetworkssynthetic, Peng_2018, chen2022understandingdomainrandomizationsimtoreal} or improved simulator fidelity to more closely mirror physical environments \cite{9811651, torne2024reconcilingrealitysimulationrealtosimtoreal, li24simpler}.  More recently, hybrid co-training strategies have emerged that blend large-scale simulated experience with real-world demonstrations in a unified learning framework \cite{nasiriany2024robocasalargescalesimulationeveryday, maddukuri2025simandrealcotrainingsimplerecipe}. Our key insight is that many challenging robot generalization axes—such as variations in camera viewpoint, lighting, and clutter—can be addressed by a cheaper, diversity-first simulation strategy that places less emphasis on physics fidelity.
\\
\noindent \textbf{Transfer from Diverse Domains.}
Beyond the specific sim-to-real problem, modern robotics leverages data aggregated from diverse, heterogeneous sources, including different hardware, camera perspectives, control interfaces, and environments~\cite{yu2017preparing, chen2019hardware, pmlr-v180-you22a, hu2022know, salhotra2023bridging, yang2023polybot, wang2024crossembodiment}. As exhaustive data collection is often infeasible due to combinatorial complexity, research focuses on transferring knowledge or adapting policies across these domains. Key techniques include domain adaptation for feature alignment~\cite{ganin2016domainadversarial, bousmalis2017using, gupta2017learning, fang2018multitask, kim2020learning, zhu2020unpaired, you2022crossdomain}, generative models for domain translation~\cite{zhang2021cycleconsistency} or targeted data augmentation such as visual inpainting to handle distractors~\cite{yu2023scalingrobotlearningsemantically} or synthesizing novel camera perspectives~\cite{chen2024mirage, tian2025viewinvariantpolicylearningzeroshot}, learning unified representations for observations/actions~\cite{martin2019variable, attarian2023geometry, wang2024crossembodiment}, and contrastive learning. A notable example of the latter is Time-Contrastive Networks (TCN)~\cite{sermanet2018tcn}, which align temporally similar trajectory segments across domains. Our work also employs contrastive learning but focuses differently, associating viewpoints from near-identical underlying states to enable finer-grained state alignment across domains. \\
\\
\noindent \textbf{Generalist Robotic Datasets and Policies.} With the increasing availability of large-scale open-source robotic datasets \cite{mandlekar2018roboturk, dasari2020robonet, fang2023rh20t, embodimentcollaboration2023open, khazatsky2024droid}, the performance of generalist robotic policies for everyday tasks has significantly improved \cite{ black2024pi0visionlanguageactionflowmodel, geminiroboticsteam2025geminiroboticsbringingai, brohan2022rt1, driess2023palme, brohan2023rt2,octo_2023}. The diversity of large-scale robotic data collection platforms and the need for different robots to perform various tasks have highlighted a key issue: determining the best way to leverage data across these platforms. Although large open-source models have benefited from utilizing data from various embodiments and domains \cite{black2024pi0visionlanguageactionflowmodel, reed2022generalist, shah2023gnm, bousmalis2023robocat, embodimentcollaboration2023open, octo_2023, driess2023palme}, the precise nature of the generalization benefits provided by these datasets remains unclear. In this paper, we explore how datasets should be collected to enable large-scale robot policies to generalize across three specific axes of diversity: lighting, distractors, and camera perspective. 

\section{Invariance Co-training}
Our objective is to develop a robot policy $\pi$ that is invariant across diverse observational perturbations, including variations in camera perspective ($E_{\text{cam}}$), lighting ($E_{\text{light}}$), and the presence of distractor objects ($E_{\text{clutter}}$). Our approach is motivated by the key insight that robust generalization stems from learning representations that are \emph{invariant to such observational perturbations ($E$) given a specific underlying state $s$ and goal $g$}. Assume a robot is solving a task by interacting with an environment defined by a state space $\mathcal{S}$, a goal space $\mathcal{G}$, and action space $\mathcal{A}$. Given a state $s \in \mathcal{S}$, goal $g \in \mathcal{G}$, the robot receives an observation $o(s; E)$, where the observational configuration $E = \{E_{\text{cam}}, E_{\text{light}}, E_{\text{clutter}}\}$ captures the camera extrinsics, lighting conditions, and scene clutter, respectively. The robot must learn a policy $a \sim \pi(\cdot \mid o(s; E), g)$ that successfully solves the task despite variation in any subset of these factors.

To solve this problem efficiently, we leverage two complementary sources of supervision: $\textit{robot demonstrations}$ and $\textit{static videos}$. Static videos are defined as multi-view recordings of a scene without robot movement, as illustrated in~\ref{fig:teaser}. 

For robot demonstrations, we assume access to a dataset $D_{\textit{d}} = \{\tau_1, \dots, \tau_K\}$ of teleoperated trajectories $\tau_n = \{\textit{O}^{n}_{0}, a_{0}), (\textit{O}^{n}_{1}, a_{1}), \ldots, (\textit{O}^{n}_{T}, a_{T})\}$ 
of a robot completing a task. Each observation set $\textit{O}^{n}_{t}$ for time step $t$ of trajectory $\tau_n$ contains multiple observations captured under different configurations of the observational factors:  $\textit{O}^{n}_{t} = \{o(s^{n}_{t}; E_{0}), o(s^{n}_{t}; E_{1}), \ldots, o(s^{n}_{t}; E_{k})\}$
where each $E_i = \{E^{(i)}_{\text{cam}}, E^{(i)}_{\text{light}}, E^{(i)}_{\text{clutter}}\}$ represents a distinct setting of observational conditions.
For static videos, we assume a dataset \( D_{\textit{s}} \) containing sets of observations of the same state under different environments:$D_{\textit{s}} = \left\{ 
\{o(s^{i}_{0}; E_{0}),\ o(s^{i}_{0}; E_{1}),\ \ldots,\ o(s^{i}_{0}; E_{\ell})\} \right\}_{i}.$
These samples provide multi-view or multi-condition images of the same static scene, but without associated actions or time progression. As such, they are less expensive and easier to collect at scale, yet still provide rich supervision for learning invariances across observational configurations.

\begin{center}
\begin{figure*}[t]
\includegraphics[width=\textwidth]{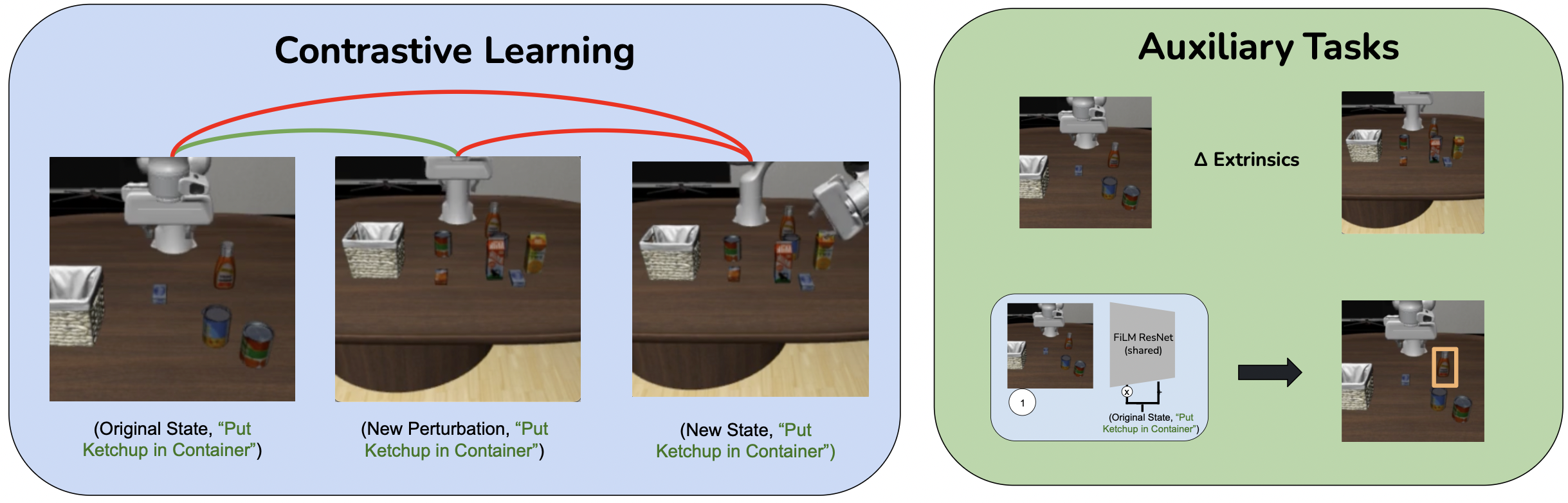}
\caption{\textbf{Our Method.} We co-train our policy with a contrastive loss using static data, two auxiliary losses using static images, and a behavior cloning loss using robot demonstration data.}
\label{fig:alignment}
\end{figure*}
\end{center}

\subsection{Decomposing Invariance Co-training}
What does a robot need to understand in order to leverage static data? We propose two key contrastive tasks to address this question: \textit{state similarity} and \textit{relative observational disentanglement}. These tasks help ensure that the policy learns representations that are invariant to semantically irrelevant factors—such as viewpoint, lighting, and clutter—while remaining sensitive to task-relevant state changes specified by the goal.

Let $(s_i, g_i)$ and $(s_j, g_j)$ be two (state, goal) pairs from possibly different trajectories, and let $E_k$ and $E_l$ denote two distinct observational configurations (e.g., different camera viewpoints or lighting setups). Let $m, n$ denote trajectory indices and $i, j$ be timesteps within those trajectories. Given a reference observation $o(s^{m}_{i}; E_k)$ with goal $g_m$, the policy must distinguish between the following scenarios, as illustrated in~\ref{fig:alignment}:

\begin{enumerate}
    \item \textbf{Same state and same goal, different observations:} $o(s^{m}_{i}; E_l)$ is visually different from but semantically equivalent to $o(s^{m}_{i}; E_k)$ under the same goal $g_m$. These observations while different should be mapped to similar representations.
    \item \textbf{Different state or different goal, same observations:} $o(s^{n}_{j}; E_k)$ may be visually similar to $o(s^{m}_{i}; E_k)$, but corresponds either to a different underlying state or a different task goal. These observations should be mapped apart.
    \item \textbf{Different state or different goals, different observations:} $o(s^{n}_{j}; E_l)$ differs in both underlying state and observational conditions (and possibly goal) from $o(s^{m}_{i}; E_k)$, and should also be mapped apart.
\end{enumerate}

We propose training a policy that learns \textit{invariant representations} under observational perturbations while maintaining sensitivity to state differences and task semantics.

\subsection{Our Objective}

We design a contrastive learning objective to encourage the policy’s vision encoder to align representations of semantically similar states—regardless of differences in observational conditions—while separating representations of distinct states. We define a total alignment loss:
\[
L_{\text{alignment}} = L_{\text{pos}} + L_{\text{neg}},
\]
where $L_{\text{pos}}$ pulls together positive pairs (same state under different observations), and $L_{\text{neg}}$ pushes apart negative pairs (different states, even if they appear visually similar).

The positive alignment loss is given by:
\begin{align}
L_{\text{pos}}(\mathcal{D}) &= \sum_{(s^{m}_{i}, s^{n}_{j}) \in \mathcal{D}^{+}} \left\| f_{\theta}(\pi_{v}(o(s^{m}_{i}; E_k), g)) \right. \nonumber \\
&\quad \left. - f_{\theta}(\pi_{v}(o(s^{n}_{j}; E_l), g)) \right\|_2^2,
\end{align}
and the negative alignment loss uses a hinge formulation with margin $\delta = 0.5$:
\begin{align}
L_{\text{neg}}(\mathcal{D}) &= \sum_{(s^{m}_{i}, s^{n}_{j}) \in \mathcal{D}^{-}} \max\bigg(0, \delta \nonumber \\
&\quad - \left\| f_{\theta}(\pi_{v}(o(s^{m}_{i}; E_k), g)) \right. \nonumber \\
&\quad \left. - f_{\theta}(\pi_{v}(o(s^{n}_{j}; E_l), g)) \right\|_2^2 \bigg).
\end{align}
Here, $\pi_v$ is the vision encoder, and $f_\theta$ is a lightweight MLP head applied on top of it. This setup enables learning a compact representation space where semantically similar states (even under differing observational conditions) are clustered together, while dissimilar states remain separable.

We formally define the construction of positive and negative pairs as follows:
\[
(s^{m}_{i}, s^{n}_{j}) \in 
\begin{cases} 
D_d^{+} & \text{if } n = m \text{ and } |i - j| < \epsilon, \\
D_d^{-} & \text{if } n \neq m \text{ or } |i - j| \geq \epsilon, \\
D_s^{+} & \text{if } n = m, \\
D_s^{-} & \text{if } n \neq m.
\end{cases}
\]
\ref{fig:alignment} illustrates example positive and negative pairs for both robot demonstration and static images.

\subsection{Auxiliary Losses for Cross-Domain Transfer}
In addition to contrastive alignment, we introduce auxiliary tasks that provide low-dimensional supervision during training time to allow the vision encoder to better learn state invariances across domains.

\paragraph{Camera Perspective: Extrinsics Regression.} 
To enhance robustness against varying camera viewpoints, we train with an extrinsics regression loss that leverages known camera transformations. This supervision helps the policy understand state invariances, recognizing when visual changes are due to perspective shifts rather than changes in the underlying robot state.

Given two observations of the same robot state $s_i$ under configurations $E_k$ and $E_l$, we define the extrinsics regression loss:
\begin{align}
L_{\text{ext}} &= \sum_{(s_i, E_k, E_l) \in D_d} \left\| g_\theta(\pi_v(o(s_i; E_k), g)) \right. \nonumber \\
&\quad \left. - g_\theta(\pi_v(o(s_i; E_l), g)) - T(E_k^{-1}E_l) \right\|_F^2,
\end{align}
where $\pi_v$ is the vision encoder, $g_\theta$ is a projection MLP, and $T(\cdot)$ is a normalization function that scales the translational component of the relative pose to address scale ambiguity.

We compute the normalization factor $Z$ for a given scene using a subset of its extrinsics ($X$):
\[
Z = \frac{1}{n} \sum_{X_k \sim D} \left\| X_k^T - \frac{1}{n} \sum_{X_l \sim D} X_l^T \right\|_F.
\]
\paragraph{Distractor Objects: Bounding Box Regression.} 
To encourage the model to focus on object-specific features and learn state-invariant representations, we use object bounding boxes as weak supervision during pretraining. Understanding object location helps the model disentangle task-relevant objects from background clutter and better recognize the same object across different viewpoints or lighting. Specifically, we supervise the encoder to predict the 2D positions of salient objects in the scene using a regression head. The loss takes the form:
\[
L_{\text{bbox}} = \sum_{(s_i, E_k) \in D} \left\| h_\theta(\pi_v(o(s_i; E_k), g)) - \text{bbox}(s_i; E_k) \right\|_2^2,
\]
where $h_\theta$ is a regression head and $\text{bbox}(s_i; E_k)$ denotes the ground-truth or predicted bounding boxes for objects in the scene.

\subsection{Putting It All Together}

We train our policy end-to-end by co-optimizing the primary behavior cloning objective with the auxiliary tasks described above. These auxiliary losses—contrastive alignment, extrinsics regression, and (optionally) bounding box prediction—encourage the vision encoder to learn representations that are invariant to observational variation while preserving task-relevant distinctions.

The total co-training loss is defined as:
\[
\mathcal{L}_{\text{total}} = \mathcal{L}_{\text{BC}} + \lambda_{\text{align}} \cdot \mathcal{L}_{\text{alignment}} + \lambda_{\text{ext}} \cdot \mathcal{L}_{\text{ext}} + \lambda_{\text{bbox}} \cdot \mathcal{L}_{\text{bbox}},
\]
where $\mathcal{L}_{\text{BC}}$ is the behavior cloning loss:
\[
\mathcal{L}_{\text{BC}} = - \sum_{(o(s_i; E_k), a_i) \in D_d} \log \pi(a_i \mid o(s_i; E_k), g),
\]
and the remaining terms are auxiliary losses described in previous sections. Coefficients $\lambda_{\text{align}}$, $\lambda_{\text{ext}}$, and $\lambda_{\text{bbox}}$ control the relative contribution of each auxiliary signal during training.

\subsection{Policy Architecture}
Inspired by scalable robot learning architectures in previous works \cite{brohan2022rt1}, we parameterize our policy with a EfficientNet-b0 ResNet encoder \cite{tan2020efficientnetrethinkingmodelscaling}, and transformer backbone. In addition, we use a diffusion policy~\cite{chi2023diffusion} action head to allow our policy to handle noise from its demonstrations as well as different strategies for accomplishing its task. We use a delta Cartesian action space with $3$ translation dimensions, $3$ Euler angle rotations, and a gripper movement.

\section{Experimental Setup}
\label{sec:result}
We aim to test the following questions:
\begin{enumerate}
    \item Can co-training with diverse, static visual data from \textbf{simulation} improve the generalization of robot policies to unseen observational configurations such as novel viewpoints, lighting, and clutter?
        
    \item Can invariance co-training unlock generalization properties from large-scale datasets without requiring additional interactive robot data?

    \item What is the impact on auxiliary tasks on the ability of our method to learn from simulation data?
        
    \item What are the effects of dataset \textbf{scale and diversity} (e.g., number of perspectives or static frames) on generalization to out-of-distribution conditions?
\end{enumerate}

\noindent \textbf{Simulation Setup.} We collect a static dataset in LIBERO, Unreal Engine, and Simpler Environment. In LIBERO and the Simpler Environment~\cite{li24simpler}, we freeze 1000 states per task mid-execution and capture 10 viewpoint images per state. In Unreal Engine, we collect a dataset consisting of $10$ different scenes, and $100$ objects per scene. Further simulation details are provided in the appendix. \\
\\
\noindent\textbf{Real-World Robotic Setup.}  
We use both a $7$-DoF Franka Emika Panda robot and a WidowX250s from the BRIDGE setup~\cite{ebert2021bridge} controlled in joint space. Two RGB cameras are mounted around the workspace to collect $128 \times 128$ image observations. We study the \textit{pen-in-cup} task on the Franka and the \textit{open faucet}, and \textit{kitchen rack} tasks on the BRIDGE toy kitchen setup. For each task, we collect 200 human teleoperated demonstrations using an Oculus headset. To induce viewpoint diversity, camera positions are changed every 10 demonstrations. We also collect a static video dataset of 50 in-domain scenes, each consisting of roughly 200 images. For each scene, the robot’s end-effector is manually moved to different poses to mimic realistic trajectories and diversify the static dataset.\\
\\
\noindent\textbf{Methodology.}  
We evaluate four variants of our method and compare them against three baselines. The baselines are: standard behavior cloning (\textbf{BC}), behavior cloning with a pretrained DINO encoder (\textbf{DINO + BC}), and behavior cloning with generative data augmentation (\textbf{BC + GenAug}), which applies standard image-space augmentations during training. For generative augmentation, we additionally compare with methods that use \textbf{ZeroNVS}~\cite{sargent2024zeronvszeroshot360degreeview} for viewpoint synthesis and image editing models (\textbf{InstructPix2Pix}~\cite{brooks2023instructpix2pixlearningfollowimage}) to simulate lighting changes and introduce distractor objects. Each method is evaluated $20$ times in the real world across uniformly-selected degrees of perturbation. Further information about evaluation is provided in the Appendix.

The methods compared include:
\begin{itemize}
    \item \textbf{Demos + Generative:} Uses robot demonstrations augmented with data from generative models for co-training.
    \item \textbf{Ours:} Co-trains the vision encoder using invariance objectives on robot demonstrations and synthetic static images.
    \item \textbf{Ours + Real:} Co-trains the vision encoder using invariance objectives on robot demonstrations, synthetic static data, and real-world static images.
\end{itemize}

\section{Experimental Results}
\begin{figure*}[t]
\includegraphics[width=\textwidth]{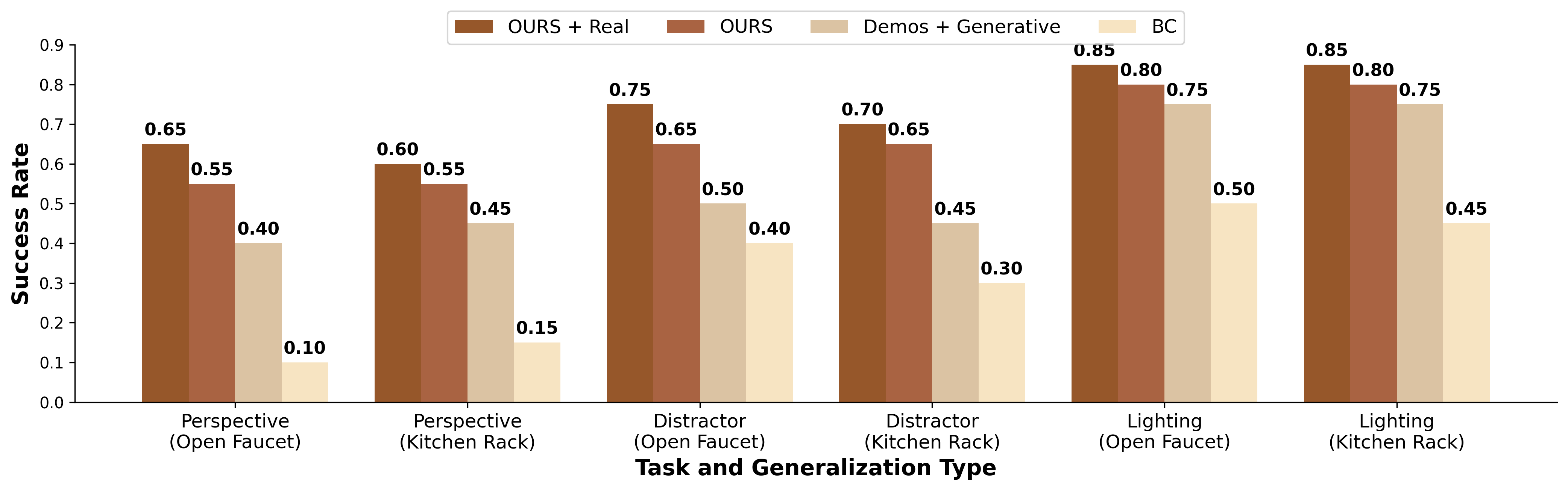}
    \caption{\textbf{Generalization Results Across Observation Variations.} Invariance Co-training achieves an average of $40\%$ higher success than standard behavior cloning across environments with perspective, distractor, and lighting variations.}
    \label{plot:alignment}
\end{figure*}

\begin{figure*}[t]
    \centering
    \begin{minipage}[t]{0.48\textwidth}
        \centering
        \includegraphics[width=\textwidth]{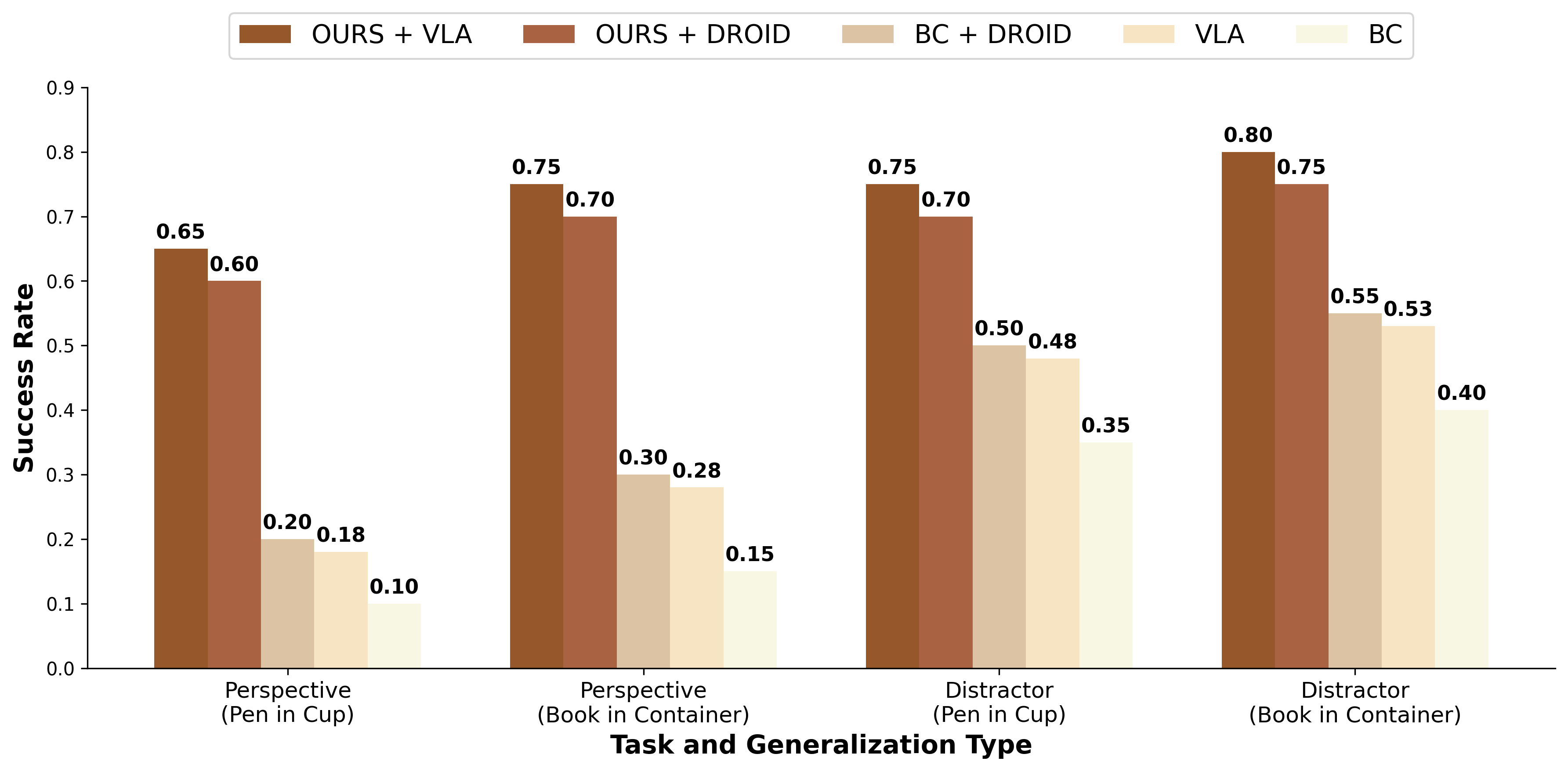}
        \caption{\textbf{Ablation: Generalization.} Invariance co-training achieves an average success rate increase of $30\%$ compared to standard co-training when leveraging the DROID dataset.}
        \label{fig:ablation_gen}
    \end{minipage}%
    \hfill
    \begin{minipage}[t]{0.48\textwidth}
        \centering
        \includegraphics[width=\textwidth]{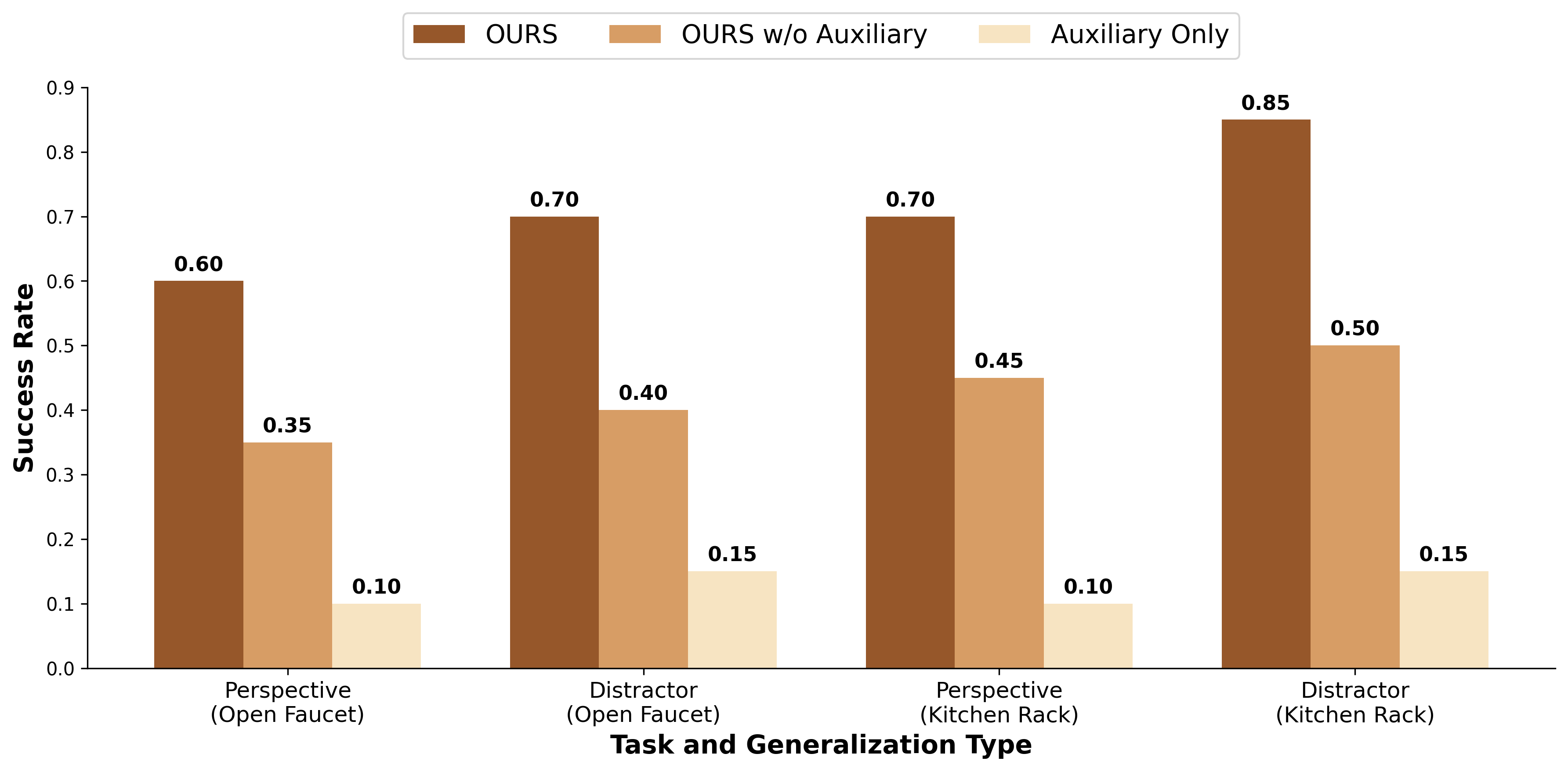}
        \caption{\textbf{Ablation: Auxiliary Tasks.} Including auxiliary tasks yields an average absolute success rate increase of approximately $29\%$ compared to omitting them.}
        \label{fig:auxiliary}
    \end{minipage}
\end{figure*}

\noindent \textbf{Our method effectively leverages robot demonstrations and static scenes to train policies that generalize across observation variations.}
Figure~\ref{fig:alignment} shows our full approach significantly outperforms baseline Behavioral Cloning (\textit{BC}), improving average success rates by approximately $40\%$ across perspective, distractor, and lighting variations. Notably, the inclusion of real static images specifically contributes consistent improvements, boosting success rates by approximately $7\%$ on average compared to variants relying only on demonstrations and simulation, highlighting the distinct value of real-world visual diversity. Furthermore, compared to augmenting demonstrations with generative models (\textit{Demos + Generative}), our method using real static images achieves superior generalization, yielding over $18\%$ higher success rates on average across these same variation categories. This shows that real static scenes can provide a direct source of diverse, realistic observations that can cover generalization gaps that generative models may struggle to synthesize accurately.
For example, while the generative method successfully generated novel camera perspectives around a $30^\circ$ rotation, it failed to generate viewpoints corresponding to more extreme perspective shifts.\\
\\
\textbf{Our invariance co-training method effectively leverages large-scale static datasets like DROID to achieve superior generalization compared to standard co-training approaches, without requiring additional robot interaction data.}
As shown in Figure~\ref{fig:ablation_gen}(a), standard co-training (\texttt{BC + DROID}) offers limited improvement over \texttt{BC} alone ($\sim$$0.20$-$0.30$ perspective success).In contrast, our invariance co-training (\texttt{OURS + DROID}) demonstrates significantly stronger generalization using the same \texttt{DROID} dataset. Compared to \texttt{BC + DROID}, our method increases perspective success rates by $0.40$ (reaching $0.60$ for Pen-in-Cup and $0.70$ for Book-in-Container) and enhances distractor robustness by $0.20$ (reaching $0.70$ for Pen and $0.75$ for Book). These results demonstrate that invariance co-training can lead to substantial gains in policy generalization large, diverse datasets, without the dataset needing to contain the same coverage of observation perturbations. \\
\\
\textbf{Auxiliary tasks are crucial for the performance of our co-training method, providing significant improvements in generalization across all tested conditions.} As evaluated in Figure~\ref{fig:auxiliary}(b), comparing our full method (\texttt{Ours with Auxiliary}) to an ablation without auxiliary losses (\texttt{OURS}), we find these losses provide substantial performance gains. On average, including auxiliary tasks increases perspective success rates by $0.25$ and distractor robustness by $0.325$, leading to an overall average success rate improvement of approximately $29\%$. While using only auxiliary losses performs poorly (\texttt{Auxiliary Only} baseline success $\sim$$0.10$-$0.15$), their integration into co-training is crucial for enhancing policy generalization. This suggests that the information the network gets from lower-dimensional state supervision is important to encourage the network to better learn invariant features across domains.\\

\begin{figure*}[t]
\includegraphics[width=0.9\textwidth]{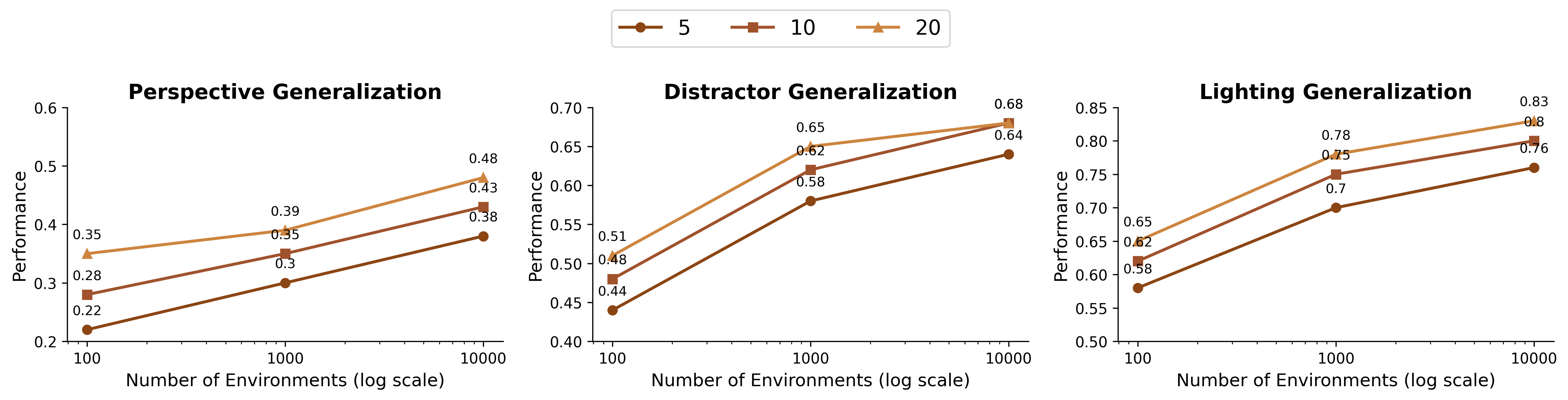}
\caption{\textbf{Ablation: Data Scaling and Diversity.} Policy generalization performance improves with dataset scale (number of environments) and diversity (number of augmentations, referred to as perturbations per state).}
\label{fig:scaling_results}
\end{figure*}

\noindent \textbf{Generalization performance demonstrably improves with both the scale and diversity of the static dataset, as evidenced by our ablation results.}
Figure~\ref{fig:scaling_results} shows performance scaling with static environment count (scale) and augmentation diversity. Increasing scale yields significant gains; perspective success with 10 augmentations, for example, rises from $0.27$ (100 envs) to $0.43$ (10k envs). Increasing the number of perturbations per state also provides consistent benefits, with perspective success at 10k environments improving from $0.38$ (5 augs) to $0.48$ (20 augs). These positive trends across perspective, lighting, and distractor variations highlight the value of large, diverse static datasets for robustness. While more perturbations per state consistently yield performance benefits across variations, the notably log-linear scaling for perspective generalization suggests this represents a particularly challenging axis of variation, potentially requiring more data diversity to master compared to lighting or distractors.
\section{Conclusion and Limitations}
\label{sec:conclusion}
\textbf{Conclusion and Future Work}
We present \textit{invariance co-training}, a method for improving robotic policy robustness to observational variations by leveraging diverse static or quasi-static visual data from simulation and the real world. Training vision encoders with auxiliary objectives designed to learn these invariances leads to substantial performance gains. Our method demonstrates significantly improved generalization, achieving an average of $40\%$ higher success than behavior cloning on large-scale datasets under challenging visual variations. We believe this work represents an important step towards thinking more closely about achieving sufficient data coverage across specific, challenging axes of generalization, rather than solely relying on scaling undifferentiated datasets. A promising avenue for future research is scaling invariance co-training across diverse tasks with vision-language-action models (VLAs) to improve their robustness across open-world environments.\\
\\

\noindent \textbf{Limitations}
While promising, our work has several limitations. Our method directly addresses robustness to viewpoint, lighting, and distractor variations; however, its ability to handle significant changes in observation, such as generalizing to entirely novel objects was not the focus and may require further investigation. Additionally, generating or collecting sufficiently diverse and potentially calibrated static/quasi-static data remains a prerequisite, though we argue this is often more scalable than collecting equivalent interactive demonstration data.


\section*{ACKNOWLEDGMENT}

The authors would like to thank the reviewers for their valuable comments and suggestions. This work was supported by [funding agency].


\bibliographystyle{IEEEtran}
\bibliography{example}

\newpage
\section*{APPENDIX}

Additional experimental details and supplementary results can be found at the project website.

\begin{figure}[htb]
    \centering
    \includegraphics[width=\columnwidth]{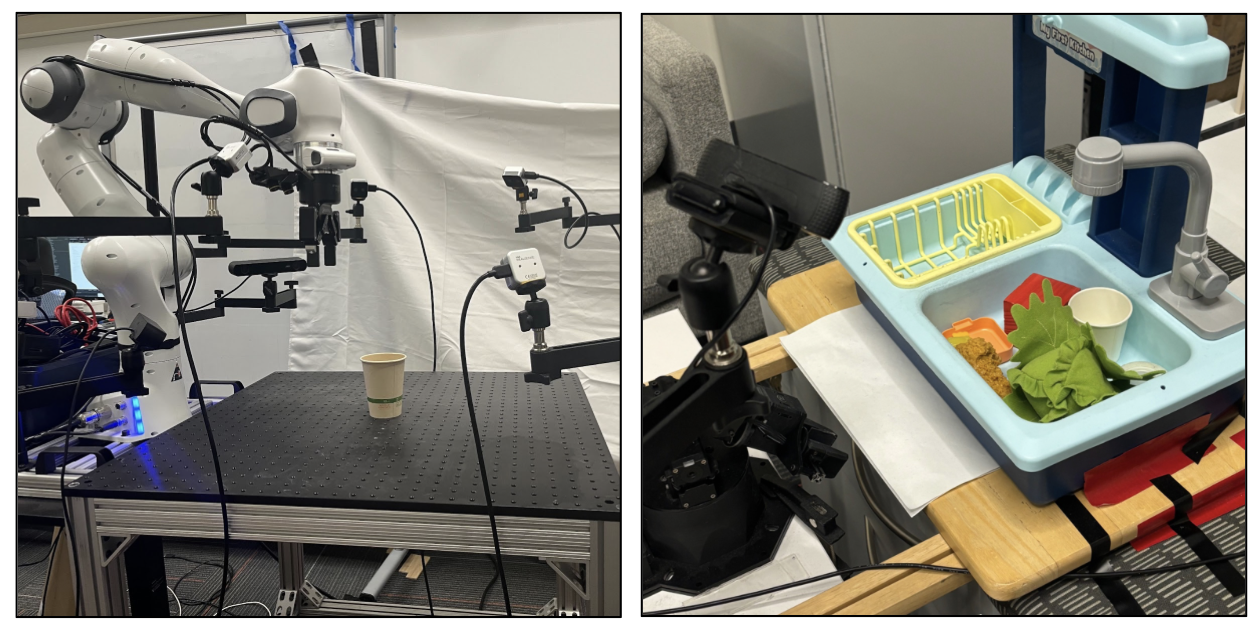}
    \caption{\textbf{Real-World Robotic Setup.} }
    \label{fig:realsetup}
\end{figure}
\label{appendix:architecture}
\subsection{Additional Real World Setup Details}
We use a modified version of the BRIDGE controller~\cite{ebert2021bridge} and integrate components from the DROID~\cite{khazatsky2024droid} codebase for low-level robot control. The controller operates at 10 Hz and commands the robot in delta Cartesian space with a 7-dimensional action space: the first three dimensions control translation, the next three control rotation (as Euler angles), and the final dimension controls the gripper.

The real-world setup includes 6 RGB cameras mounted around the scene, which are used both during data collection and evaluation. These multi-view observations enable us to simulate quasistatic visual perturbations and support the invariance objectives during training. An illustration of our setup is provided in Figure~\ref{fig:realsetup}, and additional implementation details regarding the camera synchronization infrastructure are provided in Appendix~\ref{appendix:cameraserver}.

\subsection{Camera Server}
\label{appendix:cameraserver}
In order to allow for better synchronicity, we created custom camera server to feed observations into the policy. Instead of directly reading images from each camera's feed using OpenCV Capture, this server opened several camera feeds in different threads. This multithreaded approach allowed each camera to operate independently, reducing latency and ensuring that frame capture was not bottlenecked by sequential processing as the number of cameras scaled. Each camera streams information with a framerate of 30 FPS . When the environment needs an observation it only pulls the most recent frame stored in the server, which ensures that the policy will be able to operate at a more stable framerate. In addition, care had to be taken to unsure that the cameras were routed through USB 3.0 ports, and that the computer processing the camera scheme had a sufficiently high bandwidth. Notably, the computers recommended by the DROID setup were insufficient for processing more than $2$ cameras at a time.

\subsection{Additional DROID Training Details}
Instead of training on the entire DROID split, we filtered the DROID dataset on a subset based on relevant tasks. Specifically, since our task is in an indoors setting, we filtered for trajectories containing the "marker," "cup," "drawer," and "mug" trajectories. In addition, we also sample random trajectories from the "living room," "industrial office," and "office" locations. We train on a total of around 2000 trajectories in 5 scenes. While it is possible to scale our policy can scale to fit all of the DROID dataset, we found that filtering reduces training and allows the a smaller model to better fit the data. Finally, since the DROID dataset contains data with different levels of accuracy for extrinsics callibration, when pre-training the policy, we filter for only the demonstrations that have high-quality extrinsics data.

Our method for sampling positive and negative pairs in DROID is simple. For positive pairs, we randomly sample a filtered trajectory according to the scheme above. Then, within this trajectory, we sample a random timestep and use its corresponding camera perspectives as well as their extrinsics information as a postive pair. Note that DROID uses a fixed set of cameras per trajectory, and thus, only contains two different camera perspectives per state. As previously discussed, we believe this is a limitation in training perspective-invariant camera encoders with the DROID dataset. To sample negative pairs using DROID, we similarly upweight our sampling distribution to have $50\%$ of each training batch contain trajectories from the same scene. The rest of the batch contains random camera perspective from the rest of the scenes and trajectories.

\label{appendix:simulation}
\begin{figure*}[t]
\centering
\includegraphics[width=0.8\textwidth]{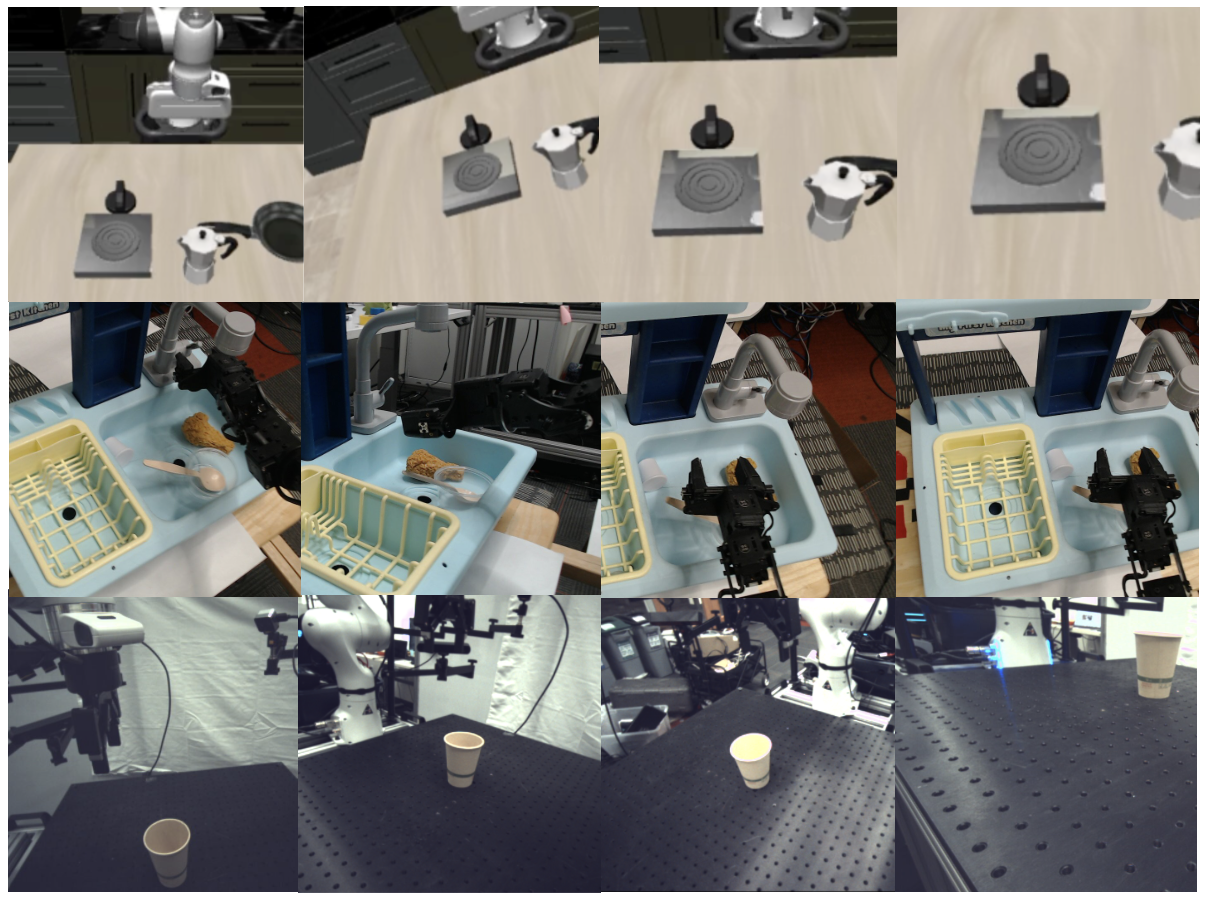}
\caption{\textbf{Evaluation Environments.} We evaluate our robot policy on LIBERO, BRIDGE, and the DROID environments.}
\label{figure:filmstrips}
\end{figure*}

\begin{figure*}[t]
\centering
\includegraphics[width=0.8\textwidth]{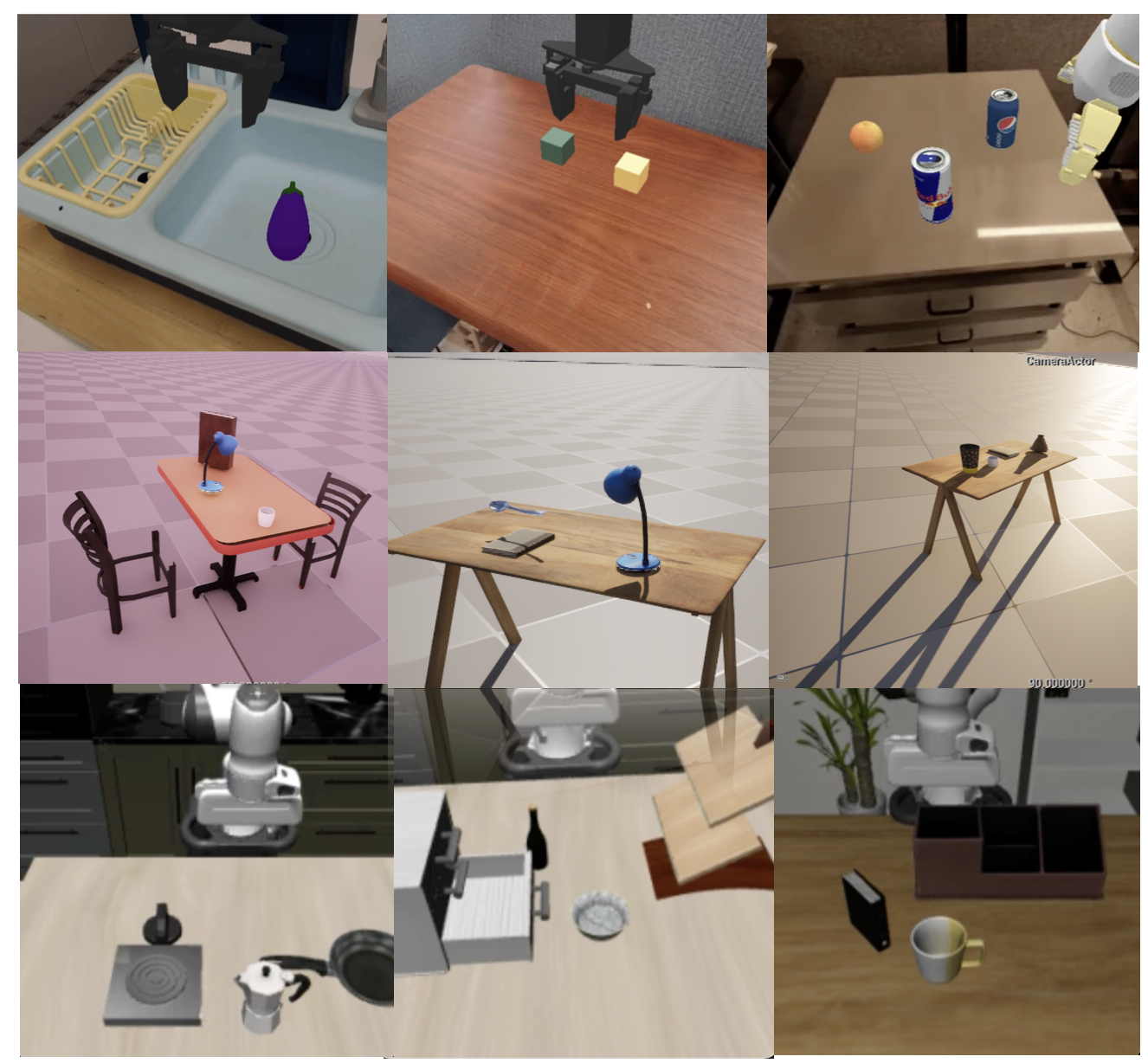}
\caption{\textbf{Simulation Examples.} Various simulation static images used to co-train our method.}
\label{figure:sim_examples}
\end{figure*}

\subsection{Additional Contrastive Training Details}
In order to sample a contrastive dataset in simulation, we roll out a trajectory in simulation and pause at a uniformly random timestep. Then, we sample $10$ different images and store these images together in our dataset. During sampling time, we randomly select a set of images, then 2 random images from this set to be our positive pair.

Contrastive losses are sensitive to the frequency at which its positive and negative comparisons are sampled. If we naively apply contrastive learning to a multitask dataset of camera perspectives, the learned network might focus more on distinguishing one dataset from another than distinguishing similar and different states. In order to encourage our vision encoder to focus on smaller difference between states within tasks, we upweight the frequency in which similar tasks are sampled. In our case, we sample half the batch as images from the same task and split the remainder of the batch as other random images.

\subsection{Policy Architecture}
We parameterize our model with a diffusion policy~\cite{chi2023diffusion}. At each state, the model passes its observation through an EfficientNet-b0 vision encoder, which outputs an aligned embedding. Note that although two vision encoders with shared weights are used during training in order to align the embeddings, only one encoder is used during evaluation time. These embeddings are then passed into a transformer, which then computes an embedding $c_{t} $that is passed as conditioning information to a diffusion head. The diffusion head consists of a noise scheduler and a noise prediction network. 

\begin{figure}[htb]
    \centering
    \includegraphics[width=0.8\columnwidth]{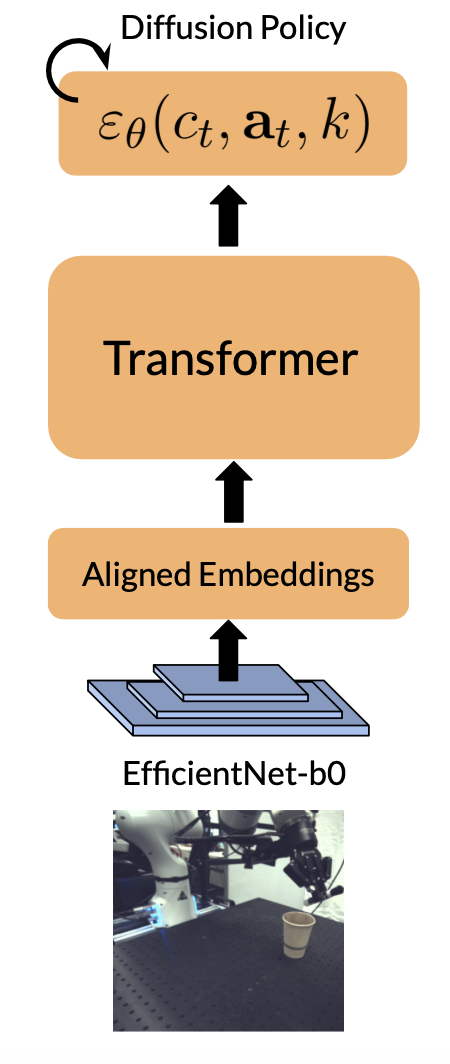}
    \caption{\textbf{Policy Architecture.} We parameterize our model with a diffusion policy.}
    \label{fig:architecture}
\end{figure}

The noise prediction network is parameterized by a U-Net found in ~ \cite{janner2022planning}. As typical in diffusion policy implementations, we simultaneously diffuse $8$ actions into the future. To obtain the actions, the model first samples a noisy action $a_{t}^{K}$ from a Gaussian distribution $\mathcal{N}(0, \sigma^{2}I)$. Then, it iteratively denoises the action using the following update rule. Figure~\ref{fig:architecture} describes the policy architecture.
$$a_{t}^{k-1} = \alpha (a_{t}^{k} - \gamma \epsilon_{\theta}(c_{t}, a_{t}^{k}, k) + \mathcal{N}(0, \sigma^{2}I))$$

During evaluation time, we sample multiple noise samples per timestep, and the average the results actions. We find that this is necessary for more dexterous manipulation tasks because otherwise, the policy would sample suboptimal actions at states which require high precision. This has a similar effect as using a deterministic (as opposed to a stochastic) robot policy.

\subsection{Model Architecture Details and Hyperparameters}
\label{appendix:hyperparams}

We use the following hyperparameters to train our policy. The top part of the table denotes hyperparameters used to train the policy, while the bottom half denotes hyperparamters for contrastive learning. 

\begin{center}
\centering
\begin{table}[h!]
\begin{tabular}{|l|c|}
\hline
\textbf{Hyperparameter} & \textbf{Value} \\
\hline
Learning Rate & 1e-4 \\
Batch Size & 128 \\
Number of Epochs & 100,000 \\
Optimizer & Adam \\
Activation Function & ReLU \\
Image Encoder Embedding Size & 256 \\
Transformer Heads & 4 \\
Transformer Layers &  4 \\
Multi-Headed Attention Feed Forward Dims & 1024 \\
Diffusion Policy U-Net Dims & 128, 256, 512 \\
\hline
\hline
Contrastive Learning Rate & 1e-4 \\
Constrastive Margin & 0.5 \\
Contrastive In-Distribution Sampling Prob & 0.5 \\
Extrinsics Regression MLP Dims&  256, 256, 256\\
\hline
\end{tabular}
\label{tab:hyperparameters}
\end{table}
\end{center}

\subsection{Additional Simulation Static Data Collection Details}
\label{appendix:sim_details}
To collect static data for invariance pretraining, we use three simulation environments: LIBERO~\cite{pumacay2024colosseumbenchmarkevaluatinggeneralization}, Simpler~\cite{li24simpler}, and Unreal Engine-based scenes. In each environment, we generate a diverse set of quasistatic observations by capturing multiple viewpoints, lighting conditions, and background distractors for fixed robot states.

In LIBERO and Simpler, we roll out scripted demonstration trajectories and pause execution at uniformly sampled mid-trajectory states. At each selected state, we freeze the robot and environment, then render observations from 10 camera viewpoints positioned uniformly along a hemisphere around the workspace. We additionally vary lighting direction and intensity across these views. This results in a total of approximately 10,000 multi-view static observation clusters per environment. All data is collected with the same camera intrinsics as those used during demonstration rollouts to ensure compatibility during co-training.

For Unreal Engine, we create 10 scene templates (e.g., kitchen, tabletops, etc.) and randomly populate each with 100 procedurally selected objects from large-scale 3D asset libraries. For each scene, we render multi-view static RGB observations with randomized lighting and clutter. Since these scenes are not bound to robot dynamics, we are able to scale the diversity of camera positions and object arrangements more easily than in physics-based simulators. These static datasets provide significant coverage over observational perturbations, and are used to train invariance objectives in conjunction with real robot demonstrations.

\subsection{Additional Simulation Evaluation Details}
\label{appendix:sim_details}

We evaluate our method in simulation on a diverse subset of tasks from the Libero-10 benchmark. These tasks involve language-conditioned manipulation, such as \textit{put both moka pots on the stove}, \textit{place the book in the back compartment of the caddy}, and \textit{put the yellow and white mug in the microwave and close it}. These scenarios require coordination across multiple objects, interaction with articulated components (e.g., drawers, cabinets), and spatial reasoning grounded in linguistic instructions. 

To isolate the impact of visual generalization, we use only the front-facing RGB camera for both training and evaluation. For certain tasks, including \textit{Open Drawer} or those with significant occlusion, we manually adjust the camera pose to ensure key objects are visible during critical interactions. This ensures that failures are attributable to generalization ability rather than camera blind spots.

Demonstrations are processed into relative Cartesian actions. For translation, we compute the positional difference between consecutive states. For rotation, we compute the relative rotation matrix and convert the result to Euler angles by multiplying the next timestep’s rotation matrix with the inverse of the current timestep’s.

\vspace{0.5em}
\label{appendix:architecture}

\noindent\textbf{Perturbation Strategy.}  
To evaluate robustness, we introduce systematic perturbations to the camera configuration and scene setup at test time. We sample camera viewpoints from three different regimes:
\begin{itemize}
    \item \textbf{30$^\circ$ rotation:} mild deviation from the demonstration perspective.
    \item \textbf{60$^\circ$ + translation:} large azimuthal shift combined with lateral camera displacement.
    \item \textbf{Uniform:} camera position sampled randomly from a hemisphere around the scene.
\end{itemize}
Additionally, we evaluate distractor robustness by inserting 1, 3, or 5 additional objects into the workspace during test time, randomly selected from Libero’s object library. Finally, we evaluate color robustness by varying lighting intensity and hue randomly per trial.

\vspace{0.5em}
\noindent\textbf{Oracle vs. Ours.}  
We compare our method against an \textbf{Oracle} policy that is trained using privileged test-time information. Specifically, the Oracle is trained with access to ground-truth camera extrinsics and an augmented dataset that includes trajectories rendered across the same perturbation regimes used during evaluation (e.g., multi-view demonstrations, distractor-rich scenes). In contrast, our method leverages only static co-training using single-view robot demonstrations and synthetic or real static images to induce invariance. This comparison quantifies how closely invariance co-training approaches the performance of a policy that is explicitly overfit to the test-time perturbations.

\begin{figure*}[t]
    \centering
    \includegraphics[width=0.9\textwidth]{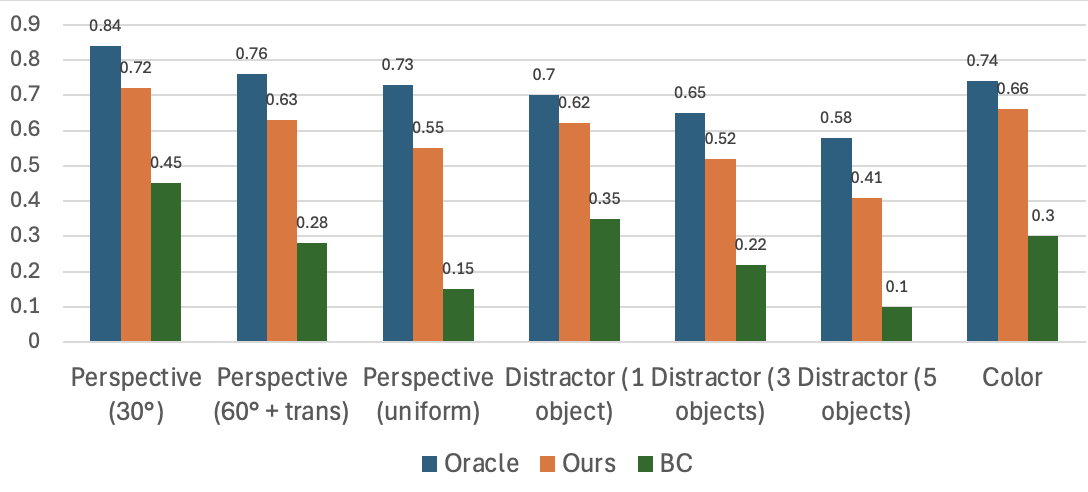}
    \caption{\textbf{Simulation Results.} Evaluation of policy success rates in Libero-10 under varied perturbation regimes. Our method outperforms behavior cloning (BC) across all axes and approaches Oracle performance, despite not using ground-truth augmentation or privileged camera supervision.}
    \label{fig:sim_results}
\end{figure*}

\vspace{0.5em}
\noindent\textbf{Analysis.}  
Simulation results are shown in Figure~\ref{fig:sim_results}. Across all perturbation types, our method significantly outperforms standard behavior cloning (BC), particularly under challenging conditions such as uniform camera perturbations and high distractor count. For example, under 60$^\circ$ viewpoint shift with translation, our method achieves a relative gain of 35 percentage points over BC. While the Oracle remains the strongest overall due to privileged access, our approach closes much of the gap—demonstrating that static co-training is sufficient to learn robust visual representations without requiring full access to perturbed data during training.

Performance also degrades gracefully as the number of distractors increases. BC deteriorates sharply (from $35\%$ with one distractor to just $10\%$ with five), whereas our method maintains over $40\%$ success even in the most cluttered scenes. This trend confirms that object-centric auxiliary supervision and contrastive co-training help the encoder focus on task-relevant regions, even in the presence of significant background noise.

Finally, under color and lighting variation, our method nearly matches Oracle-level performance, indicating strong generalization to photometric shifts—likely due to the diversity and scale of static synthetic data used during invariance training.

\section{Ablations: }

\subsection{Additional State Similarity Alignment Details}
\begin{figure*}[t]
    \centering
    \includegraphics[width=0.8\textwidth]{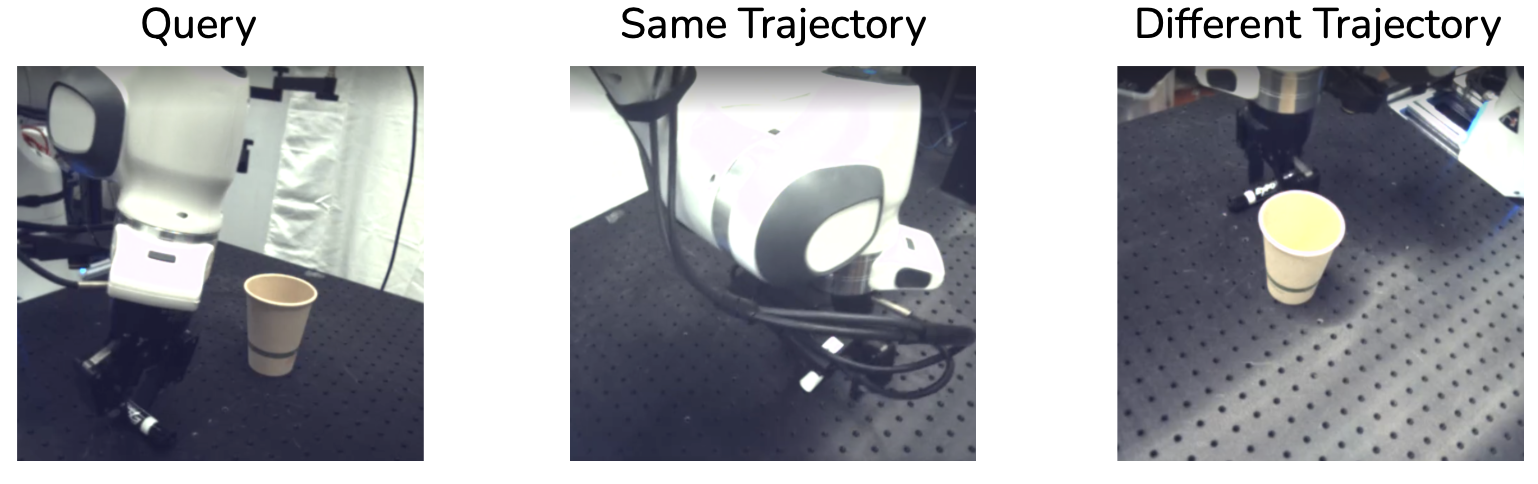}
    \caption{\textbf{Nearest Neighbor.} This figure depicts the nearest neighbors in a dataset with respect to a query image. The first dataset contains the same trajectory, while the second only contains different trajectories.}
    \label{fig:nearestneighbor}
\end{figure*}

In contrast to TCN \cite{sermanet2018tcn}, the contrastive state similarity loss aggressively maps observation together only if the state is similar enough that the policy would need to make the same decision at this state. We found that this is important to disentangling whether a change in camera perspective is from a change in state or a chance in observations. The following observation shows a nearest neighbor lookup for a vision encoder. This is done by first starting with an anchor image, then compute embeddings for all images in a validation dataset. To visualize the robustness of this approach, we computed two nearest neighbor observations--one where the dataset contains the same trajectory as the current image in a different perspective, and one where the dataset only consists of data from different trajectories. Figure~\ref{fig:nearestneighbor} shows some example results.

In the above image, the network is able to successfully find a similar image from the same trajectory and a different trajectory that is at the same state as the query image. Notably, the model doesn't only select an observation where only the pen is in a similar location, but where the end-effector is also in a similar location.

\subsection{Handheld Camera Perspective Generalization}
Humans have a remarkable ability to control understand a scene in the presence of shifting observation perspectives. However, current robot policies are not robust to this shift, causing performance to degrade significantly even when the camera is bumped slightly. 

In order to test the extreme end of this perspective generalization, we evaluate our method by moving a camera around the scene. In the real-world, this is done by a person holding the camera while the robot is doing its task. In simulation, this is done by sampling new camera angles at every state. We compute the success rate of the policy over $20$ trials. The in-distribution evaluation is done with the camera close to the scene. For out-of-distribution performance, we measure the success rate of the policy when the camera is farther away from the scene than what was collected with demonstration data.

\begin{figure}[htb]
    \centering
    \includegraphics[width=0.8\columnwidth]{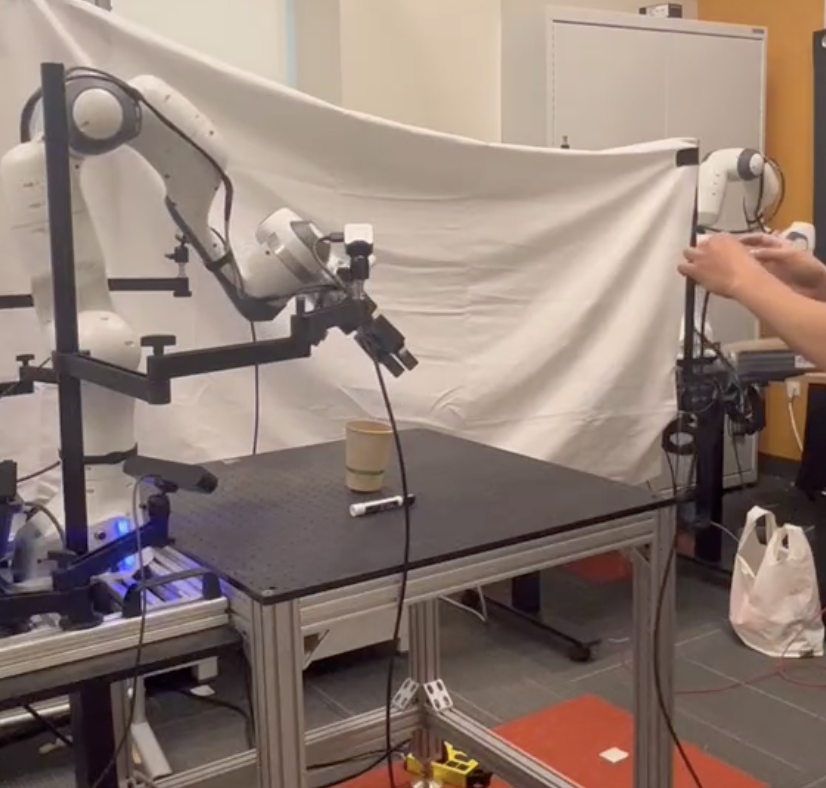}
    \caption{\textbf{Handheld Evaluation Setup.} We evaluated our robot policy with a handheld camera. For anonymity purposes, we have temporarily cut the human out of this image.}
    \label{fig:handheld}
\end{figure}

\begin{table}[!htb]
\centering
\begin{tabular}{l|c|c|c|c|c}
\toprule
& \textbf{Real} & \textbf{Close} & \textbf{Open} & \textbf{Insert} & \textbf{Reach} \\
& \textbf{World} & \textbf{Laptop} & \textbf{Drawer} & \textbf{Peg} & \textbf{Drag} \\
\midrule
Moving & 0.65 & 0.55 & 0.55 & 0.2 & 0.2 \\
Stationary (OOD) & 0.73 & 0.8 & 0.7 & 0.3 & 0.4 \\
\bottomrule
\end{tabular}
\caption{\textbf{Handheld Camera Generalization}. Vision encoder enables robot policies to complete tasks with shifting observations.}
\label{table:numcameras}
\end{table}

The results record the performance of robot policies under changing camera perspectives in both real-world and simulation settings. For comparison, we have also included the performance of stationary (out-of-distribution) tasks with our contrastive method. We find that our method achieves an average of $43\%$ success over simulation and real-world tasks, which corresponds to a $15.6\%$ drop in performance over stationary experiments. We hypothesize that the lower performance compared to static cameras can be attributed to two factors: first, the policy must continually adjust its actions to remain within the distribution of its trajectories for each camera perspective; and second, the reduced success in real-world scenarios may be due to motion blur encountered while moving the camera around the scene.

\addtolength{\textheight}{-12cm}   

\end{document}